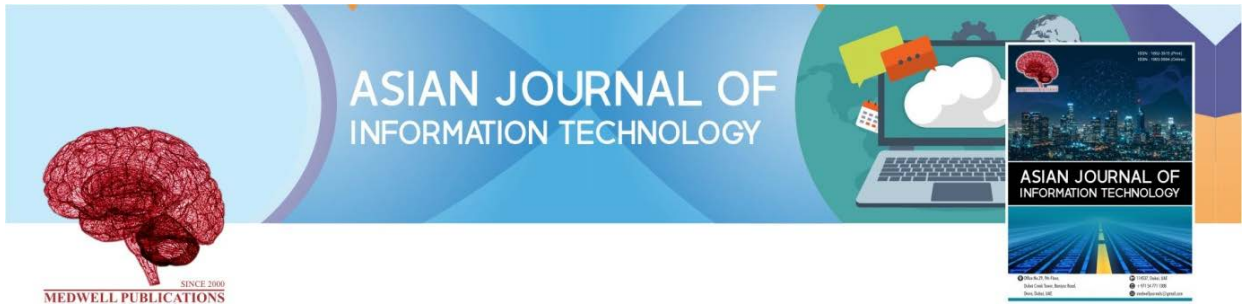

# Performance Evaluation of Classification Models for Household Income, Consumption and Expenditure Data Set

Mersha Nigus and Dorsewamy
*Department of Computer Science, Mangalore University, Karnataka, India*

**Corresponding Author:**
Mersha Nigus
*Department of Computer Science, Mangalore University, Karnataka, India*

**Abstract:** Food security is more prominent on the policy agenda today than it has been in the past, thanks to recent food shortages at both the regional and global levels as well as renewed promises from major donor countries to combat chronic hunger. One field where machine learning can be used is in the classification of household food insecurity. In this study, we establish a robust methodology to categorize whether or not a household is being food secure and food insecure by machine learning algorithms. In this study, we have used ten machine learning algorithms to classify the food security status of the Household. Gradient Boosting (GB), Random Forest (RF), Extra Tree (ET), Bagging, K-Nearest Neighbor (KNN), Decision Tree (DT), Support Vector Machine (SVM), Logistic Regression (LR), Ada Boost (AB) and Naive Bayes were the classification algorithms used throughout this study (NB). Then, we perform classification tasks from developing data set for household food security status by gathering data from HICE survey data and validating it by Domain Experts. The performance of all classifiers has better results for all performance metrics. The performance of the Random Forest and Gradient Boosting models are outstanding with a testing accuracy of 0.9997 and the other classifier such as Bagging, Decision tree, Ada Boost, Extra tree, K-nearest neighbor, Logistic Regression, SVM and Naive Bayes are scored 0.9996, 0.09996, 0.9994, 0.95675, 0.9415, 0.8915, 0.7853 and 0.7595, respectively.

## INTRODUCTION

Food insecurity remains a major development issue around the world, threatening people's well-being, growth and in some cases, survival. Efforts to address the development challenges faced by food insecurity must begin with accurate household-level measurement of key indicators. This is due to the fact that recognition of household habits related to food access serves as a vital building block for the implementation of policies and services for serving disadvantaged communities, the successful targeting of aid and the assessment of effects[1]. According to UN news, Ethiopia is a nation in the horn of Africa that is one of the fastest developing in the world. Over the last two decades, Ethiopia has experienced rapid economic growth as well as unprecedented social and human progress, a pronounced deviation from its historical trajectory. The nation has one of the world's fastest developing non-oil and non-mineral economies. According to IMF reports released in 2017, Ethiopia's economic growth has been higher than that of other African countries and it surpassed Kenya as East Africa's largest economy in 2017 (NPC's An Interim Report on Poverty Analysis Research, 2015/16).

Household expenditure surveys is used for assessing food insecurity at the national level. Food security, the opposite of food insecurity is defined as: "access by all people at all times to enough food for an active and healthy life". The most basic metric on which such steps are dependent is "household food energy supply" which is the amount of kilo-calories found in food purchased by a household[2].

As indicated by neediness examination report arranged by the WMU of the MoFED dependent on the outcomes acquired from the 1999/2000 Household Income, Consumption and Expenditure (HICE) Survey of the Central Statistical Agency (CSA), around 44% of the all out populace (45% in provincial and 37% in metropolitan zones) were discovered to be underneath destitution line while the aftereffects of the 2004/05 studies uncovered that around 39% of the absolute populace (39.3% in rustic and 35.1% in metropolitan regions) were discovered to be under the destitution line. As indicated by the report on neediness investigation study made by the MoFED utilizing the 2010/11 Household Consumption and Expenditure Survey (HCES) of the CSA shows that the destitution rate has declined to 29.6% (30.4% in rustic and 25.7% in metropolitan territories). At present, the report delivered by NPC named "Ethiopia's Progress Towards Eradicating Poverty: An Interim Report on 2015/16 Poverty Analysis Study" has put 23.5% of all out populace (25.6% in country and 14.8% in metropolitan zones) were discovered to be under neediness line in 2015/16(www.csa.gov.et).

According to Shumiye[3], Ethiopia is one of seven African countries that account for half of Sub-Saharan Africa's food poor population. According to a 1999 Food and Agricultural Organization (FAO) survey, the average caloric intake in rural areas is 1,680 kilo calories per person per day which is significantly lower than the national medically prescribed minimum daily intake of 2,100 kilo calories per person per day. According to the Federal Democratic Republic of Ethiopia Food Security Strategy (FDRE FSS) published in 1996, the required minimum daily consumption of 2,100 kilo calorie per person per day is equal to 225 kg of grain per person per year.

To our knowledge, only a few studies have used machine learning approaches to characterize household's food security status based on household revenue, consumption and spending and these studies used small data sets. In this research, we use the machine learning approach to identify the food security status of households using Ethiopian household income and expenditure data from the Ethiopian Central Statistics Agency.

**Literature review:** Any classification algorithms used household revenue, consumption and spending data in a variety of related research projects. However, the success of supervised learning in census data needs to be studied and improved. Several techniques for dealing with such massive volumes of data have been established over the years.

A research conducted by Okori and Obua[4], on the prediction of food insecurity which has yielded positive results in many parts of the world. Household food insecurity can be minimized globally if successful household food insecurity surveillance is enforced, as well as efficient implementation of government services, nutritional assistance and other policy steps.

In their analysis, [?] identified on classification models such as decision tree and random forest. They identified household characteristics that distinguish between food-insecure and food-secure households in Afghanistan, allowing for more detailed assistance targeting. According to the results of the study, income and spending goods, household size, farm-related interventions, access to specific resources and short-term shocks are all significant determinants of food security level. The model accurately classifies 80% of food-insecure households. Abdulla[5] performed a survey to classify the determinants of household food welfare, gathering primary data from sample respondents as well as secondary data from different outlets. In order to interpret the results, he used statistical statistics such as mean, standard deviation, ratio and frequency distribution. Univariate analysis such as one way ANOVA and Chi-square experiments is used to describe the features of the food safe, food insecure without hunger, food insecure with mild hunger and food insecure with extreme hunger classes. The author used 14 variables and assessed them using a logistic regression; six of them were significant at a probability level of <5% and the model correctly predicted 85.2% of the time. A study conducted by Muremyi *et al.*[6] to predict the Out of Pocket Medical Expenditures using machine learning approaches and compares the results using four machine learning approaches such as Random Forest, Decision tree Models, Gradient Boosting, Regression tree models. The data used for his analysis was collected from National Institute of Statistics (NISR) that is the Integrated Living Conditions Survey 2016-2018 (EICV5). The author used 14580 households and which represents 64314 individuals throughout the country and the information was collected at the household and the individual level. Household level information such as the out-of-pocket health expenditures such as consultation, laboratory tests, hospitalization, Diabetes, blood pressure and other illness and medication costs. From the above listed machine learning models gradient boosting was selected as the best with Train accuracy 78% and test accuracy 85%.

Author proposed[7] a method for predicting food security in Ethiopia at the sub-national level using open data linked to food insecurity drivers. The algorithm uses a random forest as the underlying algorithm and is based on an ordinal classification strategy. The model was found to be accurate to within 90% of the time.

The authors[4] used secondary data from the Uganda Bureau of Statistics for two agricultural seasons: July-December 2004 and January-June 2005, on agricultural households in the Northern, Middle, Eastern and Southern areas of Uganda (UBOS). Since, missed cases were removed, the data included 3,030 cases. To determine the impact of regional heterogeneity, data from Uganda's four regions are categorized using Support Vector Machine, k-nearest neighbor, Nave Bayes and Decision Tree. The four classification method's prediction efficiency was assessed using sensitivity, precision, accuracy and region under the curve. To see if prediction could be achieved based on regional class labels, KNN was used to apply multi label classification. Support Vector Machine outperforms the other classification algorithms with an accuracy of (96.25%).

## MATERIALS AND METHODS

**Data source:** The considered HICE information comprises of four distinctive informational indexes gathered in 2000, 2005, 2011 and 2016. The Ethiopia Household Income, utilization and expenditure (HICE) were directed by Central Statistical Agency (CSA) and National Planning Commission along with the then Ministry of Finance and Economic Development (MoFED) (CSA). This work used HICE as a data source, especially regarding household food security status to identify whether a Household is Food secure or food Insecure. we use a national Representative of 58,064 households throughout the country and out of which 32,209 instances belong to one class and 23,668 instances of another class.

**Proposed methodology:** The proposed methods classify the Food security status of household's with maximum accuracy. We use different machine learning algorithms that can help in classifying. To get improved classification accuracy, we can use more than one algorithm. Figure 1 explains the proposed work. The HICE data set is supplied to the system that is then pre-processed, so that, the data can be evaluated in a usable format. When the data set is not structured, massive, or has meaningless features, we picked the best features using feature selection. After this we use ten-fold-cross validation to split data set for both the testing and training. Then, we applied a specific machine learning algorithm to learn and apply it for testing and we also measured its performance with performance metrics of each machine learning classifier. Finally, we compared the results.

**Machine learning models:** Machine learning methods in many fields are an important method for predicting and making decisions[8]. We used the following classifiers to classify HIV test based on the research data sets, these are:

**Naive Bayes (NB):** Is a basic model of the concept of generative probabilistic classification that implies equality of entity characteristics to be categorized[9]. The Naive Bayes classifier then applies Bayes theorem assuming the existence or absence is irrelevant to certain features. Given its assumption of independence, its designation effectiveness has been proved[10]. We are using these classification methods in HICE real data set to evaluate the features.

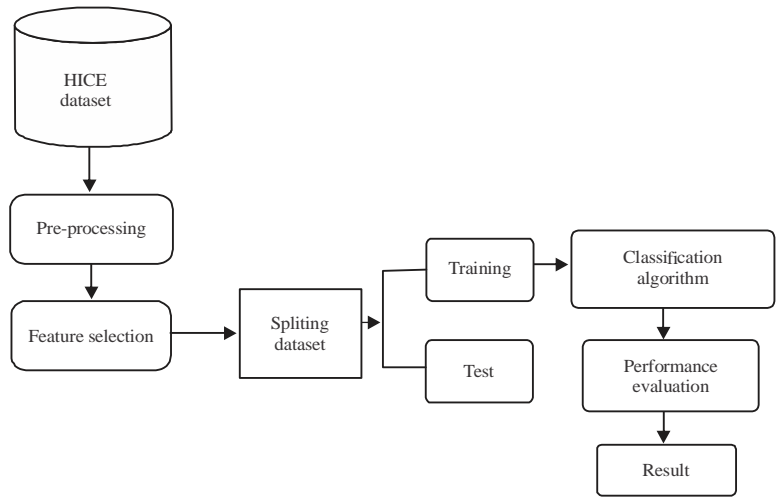

Fig. 1: Proposed methodology

**Gradient Boosting (GB):** Is an ensemble learner which means it can construct a final model based on a set of individual models. Following the development of the initial base learner, the sequence of each tree is fitted to the "pseudo residuals" of the previous tree estimation with the aim of eliminating errors[11].

**Random Forest (RF):** Is an ensemble learning methodology that employs classification decision trees. It blends various decision trees to build a final classifier. Minor variations in the trees will result from changes in each decision tree by constructing an ensemble of many uncorrelated decision trees and then combining the results. Then we'll look at a more detailed example of how to train with RF[11]. Since, the RF classifier aggregates the decisions of individual trees, it exhibits extreme generalization.

**Extra Tree (ET):** Is a way of knowing the ensemble learning based on decision trees.
Unlike Random Forest, Extra Trees Classifier randomizes all decisions and sub-sets of data to reduce both over learning and over-fitting[9].

**Bagging:** Is a meta-estimator ensemble that fits base classifiers on the preliminary data-set's random subsets and then aggregates their predictions (either by means of voting or via. averaging) to create a ultimate prediction[12].

**K-Nearest Neighbor (KNN):** Is a parameter-free model that makes use of the quantity of neighbors' k to get to the bottom of the model's complexity. Selecting random subsets of points or optimized subsets of features; after choosing features, k is commonly optimized for every subset earlier than combining the models into the ensemble[13].

**Logistic Regression (LR):** In this article, we will focus on whether the food security status of the Household is secure or insecure in numeric approach. LR evaluates a collection of data where there are one or more independent variables that decide an outcome. The dependent variables in LR are binary or dichotomous, i.e., the data is either 1 or 0. Then, the model of logistic regression takes the form.

**Decision Tree (DT):** It contains of nodes that outline checks on variables which separate data into small sub-sets and the sequence of a leaf of nodes that allocate to every corresponding segment. For our analysis, we selected the frequent DT classifier C4.5 which makes use of the precept of knowledge entropy to assemble decision trees[14].

**Support Vector Machine (SVM):** The basic concept is to build a hyper-plane in every transformed feature area, with splitting the full margin. However, it uses the kernel replacement principle to transform it into a non-linear model, instead of allowing one to describe the exact transformation.

**Ada-Boost (AB):** Performs the classification by choosing only certain distinct characteristics that can better be differentiated between the classes. It preserves the training sample's distribution range of probabilities and updates the distribution of possibility of each iteration for each test. The member classifier is developed using a specific learning algorithm and the error rate is calculated on the training data. Ada Boost uses an error rate to change the distribution of training data in probability[15].

**Performance evaluation:** All models must be evaluated using various evaluation criteria before a classification model can be defined[16]. Precision, precision, recall, f1-score and AUC/ROC are all standard parameters used to evaluate model results. In the total data-set, the precision determines the percentage of accurately calculated cases. Where data is corrupted or more instances belong to one party, consistency is often impossible to achieve. Performance Evaluation of Classification Models for HICE Dataset should also consider precision and recall. Furthermore, the f1-score is a simple harmonic mean of precision and recall magnitude. These criteria are crucial for assessing the efficacy of a classification model since they are more stable than precision. The degree of separability or scale is shown by the ROC/AUC. It demonstrates how well one model can distinguish between classes. The uncertainty matrix is often often used as a basis for assessment. The confusion matrix is critical for identifying correctly classified and incorrectly classified samples in test results. Samples are correctly classified around the diagonal but incorrectly classified around the diagonal elements.

## RESULTS AND DISCUSSION

The experimental findings of the proposed methods on the HICE data set are presented in this section. We agreed to use tenfold cross-validation in our studies. This method was chosen to illustrate how well the classifier does over a wide range of results. We used training data for HICE grouping and evaluation data for HICE training and learning. As seen in Fig. 2, the model is trained and validated using a variety of recognized classifiers (classification techniques used). We put the classification model to the test while we were developing it. To that end, we compared the accuracy metrics (training and testing) for GB, RF, KNN, ET, Bagging, DT, LR, SVM, AB and NB from different classifiers based on precision, recall, f1-score and AUC values.

| CA Name | Train Accuracy | Test Accuracy | Precision | Recall | F1-score | AUC |
|---|---|---|---|---|---|---|
| RandomForestClassifier | 1.0000 | 0.9997 | 0.999846 | 0.999537 | 0.999691 | 0.999671 |
| AdaBoostClassifier | 0.9995 | 0.9996 | 0.999537 | 0.999691 | 0.999614 | 0.999554 |
| BaggingClassifier | 1.0000 | 0.9996 | 0.999691 | 0.999537 | 0.999614 | 0.999574 |
| GradientBoostingClassifier | 0.9999 | 0.9996 | 0.999691 | 0.999537 | 0.999614 | 0.999574 |
| DecisionTreeClassifier | 1.0000 | 0.9996 | 0.999691 | 0.999537 | 0.999614 | 0.999574 |
| ExtraTreesClassifier | 1.0000 | 0.9901 | 0.990594 | 0.991664 | 0.991129 | 0.989892 |
| KNeighborsClassifier | 0.9575 | 0.9440 | 0.924163 | 0.980086 | 0.951304 | 0.939313 |
| LogisticRegressionCV | 0.8886 | 0.8890 | 0.901563 | 0.899197 | 0.900379 | 0.887671 |
| GaussianNB | 0.7895 | 0.7913 | 0.851666 | 0.757796 | 0.801993 | 0.795646 |
| LinearSVC | 0.7453 | 0.7484 | 0.924749 | 0.597561 | 0.725994 | 0.768109 |
| PassiveAggressiveClassifier | 0.6722 | 0.6751 | 0.632300 | 0.997839 | 0.774085 | 0.632902 |
| BernoulliNB | 0.5576 | 0.5578 | 0.557823 | 1.000000 | 0.716157 | 0.500000 |

Fig. 2: Model performance metrics

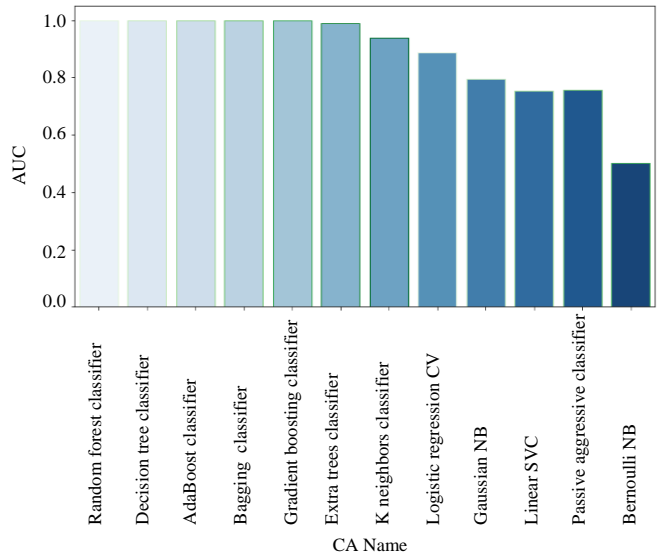

Fig. 3: Comparison of ten classifier with AUC

The best results we get using the proposed methods are those generated by the classifiers. Our aim was to determine which classification model was the most effective and explain it. A large number of trials are carried out to ensure that the proposed procedure is accurate and the results are interpreted using normal metrics such as accuracy, precision, recall, F-1 score and AUC/ROC as seen in Fig. 2. As Fig. 3 shows that, the performance of all classifiers has better results for all performance metrics. The proposed approach efficiency for the accuracy of the training value is higher than 0.80% is achieved except Gaussian-NB, Linear SVC, Passive Agressive Classifier and Bernoulli NB. The other performance metrics like recall, F1-score and AUC were also commendable. As shown in Table 1, the results of the random forest, extra tree and decision tree and bagging classifier is best with a training accuracy of 1.00 and the score of Gradient-Boosting classifier, Ada-Boost classifier, K Neighbors classifier and Logistic Regression are 0.9999, 0.9995, 0.9575 and 0.8886, respectively.

Figure 4 contrasts the performance results of the ten ML techniques using AUC/ROC. AUC/ROC curves allow the depiction of the relationship of the TPR and FPR, whereas AUC helps to evaluate different algorithms or combinations of hyper-parameters (Fig. 5 and 6).

*Asian J. Inform. Technol., 20 (5): 134-140, 2021*

Table 1: Data set description

| Attributes | Values | Mean | Count | Min | Max | SD | Description |
|---|---|---|---|---|---|---|---|
| Zone | 1-22 | 5.975338 | 58064 | 1.000000 | 22.000000 | 4.990708 | Zone |
| Woreda | 1-25 | 6.42820 | 58064 | 1.000000 | 25.00000 | 5.41742 | Woreda |
| Region | 1-7 & 12-15 | 6.248553 | 58064 | 1.000000 | 15.000000 | 4.282908 | State |
| Town | 1-4 & 8 | 3.597375 | 58064 | 1.000000 | 8.000000 | 3.302148 | Town |
| K ketema | 1-10 & 88-89 | 32.963213 | 58064 | 1.000000 | 99.000000 | 41.426997 | Sub city |
| Kebele | 1-39,401-403 & 99 | 18.098788 | 58064 | 1.000000 | 999.000000 | 102.601826 | Kebele |
| HH_SNO | 1-23 & 74 | 7.913165 | 58064 | 1.000000 | 74.000000 | 4.457474 | Household serial Number |
| HH_Size | 1-28 | 4.205153 | 58064 | 1.000000 | 28.000000 | 2.343605 | Household size |
| ADEQIV | 4.5-23.9 | 3.492198 | 58064 | 0.500000 | 23.900000 | 1.959516 | Adult Equivalent Quintile |
| UR | 1,2 | 1.356331 | 58064 | 1.000000 | 2.000000 | 0.478919 | Residence place |
| REP | 1-50 | 19.989253 | 58064 | 1.000000 | 50.000000 | 12.437010 | Report level |
| Ecology | 1,2,3,9 | 4.274990 | 58064 | 1.000000 | 9.000000 | 3.163983 | Ecology |
| Sex_Head | M,F | 1.326381 | 58064 | 1.000000 | 2.000000 | 0.468893 | Household Head Sex |
| Age_Head | Continuous | 41.630821 | 58064 | 11.000000 | 99.000000 | 15.656532 | Household Head Age |
| M$_s$tatus_head | 1,2,3,4,5,6 | 2.307781 | 58064 | 1.000000 | 6.000000 | 0.989209 | Household Head Martial Status |
| Education_Head | 1,2 | 1.396304 | 58064 | 1.000000 | 2.000000 | 0.489133 | Household Head Education |
| WGT | 7.5 -7384.5 | 609.233449 | 58064 | 7.500000 | 7384.700000 | 751.931360 | Weight |
| Annual_Expend | 86-179558 | 3527.057919 | 58064 | 86.000000 | 179558.000000 | 3615.923516 | Household Annual Expenditure |
| Year | 2011-2016 | 2013.603076 | 58064 | 2011.000000 | 2016.000000 | 2.497896 | Survey Year |
| Net_Calorie | 128-61245 | 2620.257044 | 58064 | 128.000000 | 61245.000000 | 1238.236500 | Net Calorie |
| Security_Status | 1,2 | 0.557626 | 58064 | 0.000000 | 1.000000 | 0.496672 | Food Security Status of Household |

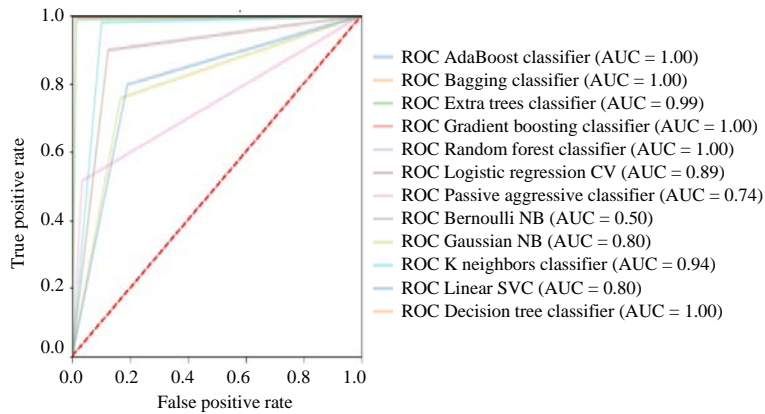

Fig. 4: Comparison of ten classifier with ROC

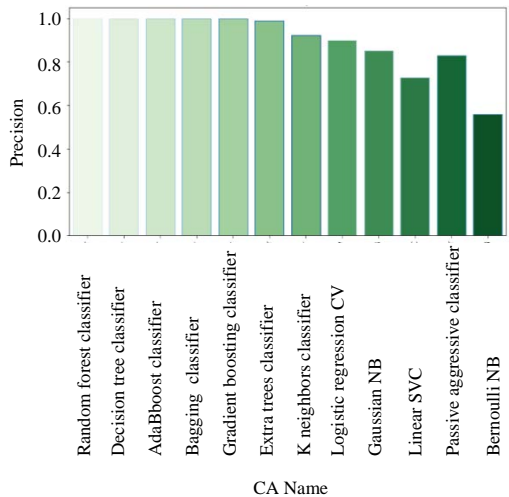

Fig. 5: Precision result of classifiers

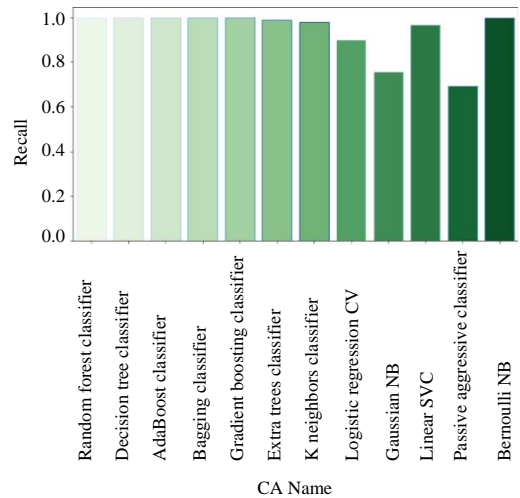

Fig. 6: Recall result of classifiers

## CONCLUSION

This study presents a proposed methodology to classify the status of household food insecurity using different machine learning algorithms For experiment purposes, we used ten classification models, namely; Random Forest, Extra Tree, Decision Tree, Bagging, K-nearest neighbor, Gradient Boosting, Logistic Regression, SVM, Ada Boost and Naive Bayes. The purpose of this work was to compare algorithms with different performance measures. The performance measures such as accuracy, recall, precision, f1-score, AUC/ROC and confusion matrix were employed to confirm the validity of the method. The experimental results of the proposed method exhibit better classification of the given dataset, i.e., HICE, by scoring training accuracy up to 1.00 and in testing accuracy, it scored up to 0.9997. The study can be considered as a pilot run and its approach may prove to be beneficial in further studies.